\documentclass{article}

\usepackage[english]{babel}

\usepackage[letterpaper,top=2cm,bottom=2cm,left=3cm,right=3cm,marginparwidth=1.75cm]{geometry}

\usepackage{amsmath}
\usepackage{graphicx}
\usepackage[colorlinks=true, allcolors=blue]{hyperref}
\usepackage{hyperref}

\usepackage{color}

\usepackage[dvipsnames]{xcolor}
\newcommand{\worker}[1]{\textcolor{orange}{\textbf{worker}}}
\newcommand{\analyzer}[1]{\textcolor{ForestGreen}{\textbf{analyzer}}}

\title{\textbf{vLLM Hook} v1: A Plug-in for Programming Model Internals on vLLM}
\author{%
\textbf{Maintainers:} \href{https://github.com/IRENEKO}{Ching-Yun Ko}, \href{https://github.com/fangyunh}{Yunhua Fang},
and
\href{https://github.com/pinyuchen}{Pin-Yu Chen}
\\
\textbf{Community Contributors:} \href{https://github.com/tburleyinfo}{Timothy Burley},
\href{https://github.com/asanth7}{Arav Santhanam},
\href{https://github.com/rishabhsinha17}{Rishabh Sinha},\\
\href{https://github.com/danishcontractor}{Danish Contractor},
\href{https://github.com/akshitsrivastava04}{Akshit Srivastava},
\href{https://github.com/lingyue404}{Ling Yue},
\href{https://github.com/Samarpit-bhatia}{Samarpit Bhatia}, and
\href{https://github.com/Songjw133}{Jinwang Song} \\
Contact Emails: \{cyko,pin-yu.chen\}@ibm.com}

\begin{document}
\maketitle

\begin{abstract}
Modern artificial intelligence (AI) models are deployed on inference engines to optimize runtime efficiency and resource allocation, particularly for transformer-based large language models (LLMs).
The vLLM project\footnote{\url{https://github.com/vllm-project/vllm}} is a major open-source library to support model serving and inference. However, the current implementation of vLLM limits programmability of the internal states of deployed models. This prevents the use of popular test-time model alignment and enhancement methods. For example, it prevents the detection of adversarial prompts based on attention patterns or the adjustment of model responses based on activation steering.
To bridge this critical gap, we present \textbf{vLLM Hook}, an opensource plug-in to enable the programming of internal states for vLLM models. Based on a configuration file specifying which internal states to capture, \textbf{vLLM Hook} provides seamless integration to vLLM and supports two essential features: \textit{passive programming} and \textit{active programming}. For passive programming, \textbf{vLLM Hook} probes the selected internal states for subsequent analysis, while keeping the model generation intact. For active programming, \textbf{vLLM Hook} enables efficient intervention of model generation by altering the selected internal states. In version 1 (named \textbf{vLLM Hook MIA}~\cite{fang2026mia}), the plugin runs on vLLM's V2 model runner and preserves the engine's CUDA graphs by default, so that capturing and steering internal states incurs only a small decoding overhead. It supports both offline use and serving through \texttt{vllm serve}.  In addition to presenting the core functions of \textbf{vLLM Hook}, we demonstrate use cases including prompt injection detection, attention-based re-ranking for retrieval-augmented generation (RAG), activation steering, hidden-state probing, hallucination detection, schema linking for text-to-SQL, and tool-call risk assessment. Finally, we welcome the community's contribution to improve \textbf{vLLM Hook} via \url{https://github.com/IBM/vLLM-Hook}.

\end{abstract}

\section{Overview of vLLM Hook}

\paragraph{What is \textbf{vLLM Hook}?} In the deployment of AI technology, service providers use inference engines to optimize the economic factors of the hosted model, such as latency, memory, energy, hardware, and other resource constraints. Disabling features that were otherwise programmable during the model development phase can significantly reduce inference costs. In particular, the vLLM project \cite{kwon2023efficient}, is a popular open-source, community-driven inference engine that supports the efficient serving of modern AI models, especially transformer-based large language models (LLMs). In short, we refer to a model deployed on vLLM as a ``vLLM model.''
Despite achieving high inference throughput, the current vLLM package does not support accessing or modifying important internal states of a vLLM model. Examples of these internal states include attentions, attention heads, and activations. The lack of programmability in significant internal states creates a technological gap that prevents the use of advanced techniques for monitoring, adjusting, and controlling deployed models at inference time. To bridge this gap, our vLLM Hook is an opensource plug-in of vLLM to enable flexible programming of internal states within vLLM models. 

\begin{figure}
    \centering
    \includegraphics[width=1\linewidth]{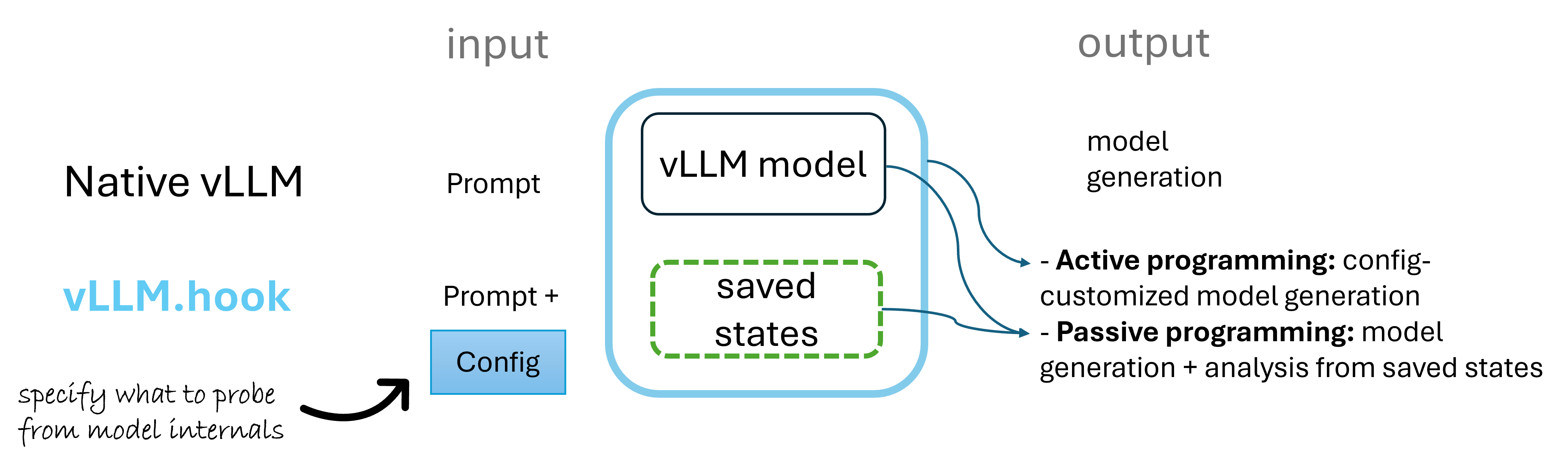}
    \caption{Overview of \textbf{vLLM Hook} (\url{https://github.com/ibm/vllm-hook}), an opensource plug-in to enable flexible programming of internal states for models deployed on vLLM. With a customizable configuration file (Config), vLLM Hook supports two essential features: (1)
    \textit{Active Programming}: \textbf{vLLM Hook} provides efficient intervention of model generation by altering the selected internal states; (2)
    \textit{Passive Programming}: \textbf{vLLM Hook} probes and saves the selected internal states for subsequent analysis, while keeping the model generation intact.  }
    \label{fig:vLLM_Hook_overview}
\end{figure}

Our \textbf{vLLM Hook} provides seamless and non-intrusive integration to vLLM. The system overview of vLLM Hook is  illustrated in Figure \ref{fig:vLLM_Hook_overview}.  
Based on a configuration file (Config) specifying which internal states within a vLLM model to ``hook'', \textbf{vLLM Hook} supports two essential features: \textit{passive programming} and \textit{active programming}. 
\begin{itemize}
    \item \textit{Passive programming} probes and saves the selected internal states for subsequent analysis, while keeping the model generation intact. 
    \item \textit{Active programming} enables efficient intervention of model generation by altering the selected internal states.
\end{itemize}

\paragraph{Why do we need \textbf{vLLM Hook}?}
Generative AI (GenAI) technology enables rapid iterations, frequent model updates, and new releases every few weeks. Once models are deployed through inference engines, derived applications and services can be rapidly built and used. However, when failure modes are identified or model operation customization requests are made, taking down the model in service for further retraining and redeployment would cause catastrophic disruption to dependent users. Therefore, there is a practical demand for GenAI service providers to implement flexible model configurations and patches ``on-the-fly'' (i.e., during deployment) for models deployed on inference engines. 

We position the vLLM Hook project as a developer kit that supports various use cases requiring the programming of vLLM model internal states. To demonstrate the programmability of different internal states through the vLLM Hook, we developed three examples that address the practical need to capture or modify specific parts of a vLLM model, including attentions, attention heads, and activations. We highlight three representative examples below, and list the remaining use cases in Section~\ref{sec:demos}.

\begin{itemize}
    \item \textbf{In-model Monitoring} [passive programming]. Use vLLM Hook to capture critical ``inference traces'' generated during inference. The saved  
    states are then used for model monitoring and misalignment detection. One example is the Attention Tracker \cite{hung2025attention}, which uses selected attention statistics from a transformer-based vLLM model to detect prompt injection attempts. Hidden states captured from selected layers can likewise feed lightweight probes for hallucination detection~\cite{cao2025detecting,yocam2026h}. The ``in-model safety guardrail'' design, enabled by vLLM Hook, differs from the typical ``cascaded safety guardrail'' design. The latter requires the use of an additional moderation model to inspect the input and output of the deployed model.

    \item \textbf{Model Steering} [active programming]. Use vLLM Hook to modify the internal states of the model to steer its output toward desired behaviors. Although recent model steering methods demonstrate promising inference-time alignment capabilities without retraining the deployed model 
    \cite{aisteer360}, the current vLLM package prevents most model steering methods that modify internal states from being implemented. One example is Activation Steering, which injects a steering vector into some layer of a deployed model for intervention, such as improving the capability in instruction following \cite{stolfo2025improving}. 
    \item \textbf{Selective Retrieval and Ranking} [passive programming]. Use vLLM Hook to activate a selected set of components within a deployed model for efficient and effective data retrieval and ranking. The rationale is to retrieve data embeddings from a deployed model that are relevant to the task of interest. One example is to only activate a selected set of attention heads within a deployed model to improve the re-ranking performance in information retrieval \cite{tran2025contrastive}. Another is to read the model's attention at the generation-start position to rank the database columns relevant to a question for text-to-SQL schema linking \cite{song2026attnlink}.
\end{itemize}

\paragraph{How do we use vLLM Hook?} We envision vLLM Hook as an organic innovation engine that will enable the development and deployment of advanced techniques for monitoring and adjusting vLLM models. As a developer kit, vLLM Hook augments the native vLLM project to support programming the internal states of vLLM models. We expect that by open-sourcing vLLM Hook, the community will be able to build valuable assets on deployed AI models for operational management and technical governance. This will also help close the gap between programming internal states during the development phase and deploying such capabilities during the deployment phase.

Finally, we list some best practices of vLLM Hook.
\begin{itemize}
    \item \textbf{Build, Probe, and Program.} An important innovation of vLLM Hook is the inclusion of the configuration file (Config) that specifies where to hook and what to program for the deployed models. Such a Config file is model-specific and task-oriented. Therefore, as illustrated in Figure \ref{fig:cycle}, the complete cycle of vLLM Hook involves three key steps:
    \begin{enumerate}
        \item \textit{Build}: In the upstream development phase, use the model to be deployed on vLLM to design the programming function and identify important internal states to hook. This build step does not require the use of vLLM.
        \item  \textit{Probe}: Based on the build results in step (1), specify the target internal states to hook by creating the Config File. 
        \item \textit{Program}: use the Config File to program the corresponding vLLM model. 
    \end{enumerate}
    
\begin{figure}[t]
        \centering
        \includegraphics[width=1\linewidth]{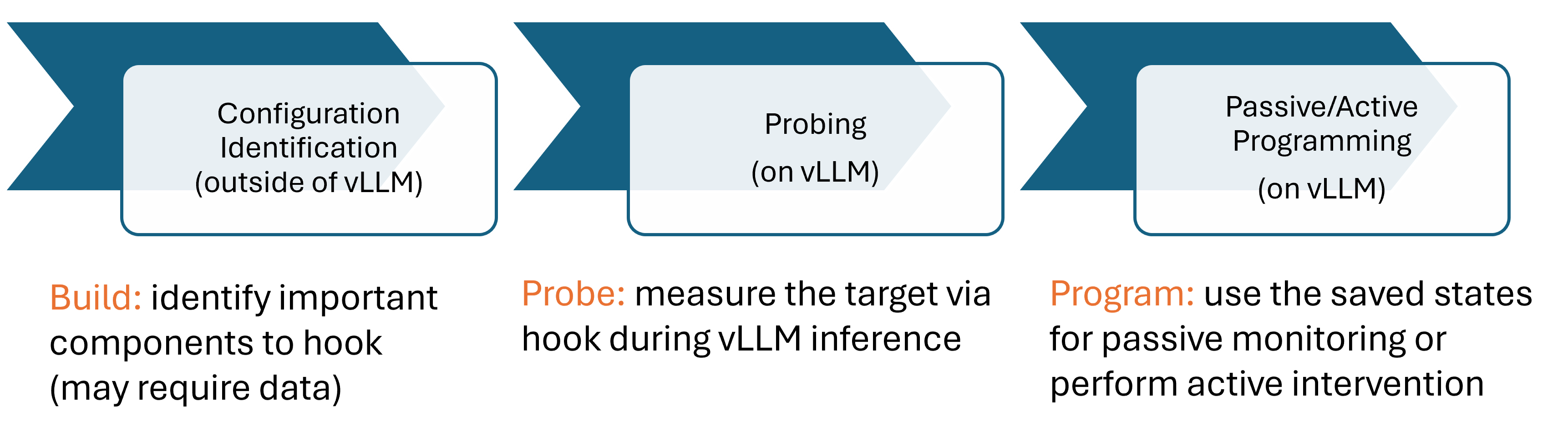}
        \caption{Complete development cycle of vLLM Hook. The three key steps are (1) \textit{Build}: In the upstream development phase, use the model to be deployed on vLLM to design the programming function and identify important internal states to hook. (2) \textit{Probe}: specify the target internal states to hook by creating the Config File. (3) \textit{Program}: use the Config File to program the corresponding vLLM model. }
        \label{fig:cycle}
    \end{figure}
        Through opensorcing and community-driven contributions, these config files can be used by different developers in their own work streams. The Build step is beyond the current scope of the vLLM Hook project. 
    \item \textbf{Be Mindful of the Progrmmability and Usability Trade-off}. Although vLLM Hook allows developers to program deployed models on vLLM, programming too many states or using computationally intensive programming functions can significantly impact the usability of inference engines. We advise developers to consider the trade-off between programmability and utility (e.g., latency of the vLLM Hook model and memory usage for programming) and to build viable configuration files with minimal hooks to internal states.
\end{itemize}

\section{Core Functions of \textbf{vLLM Hook}}

\begin{figure}
    \centering
    \includegraphics[width=1\linewidth]{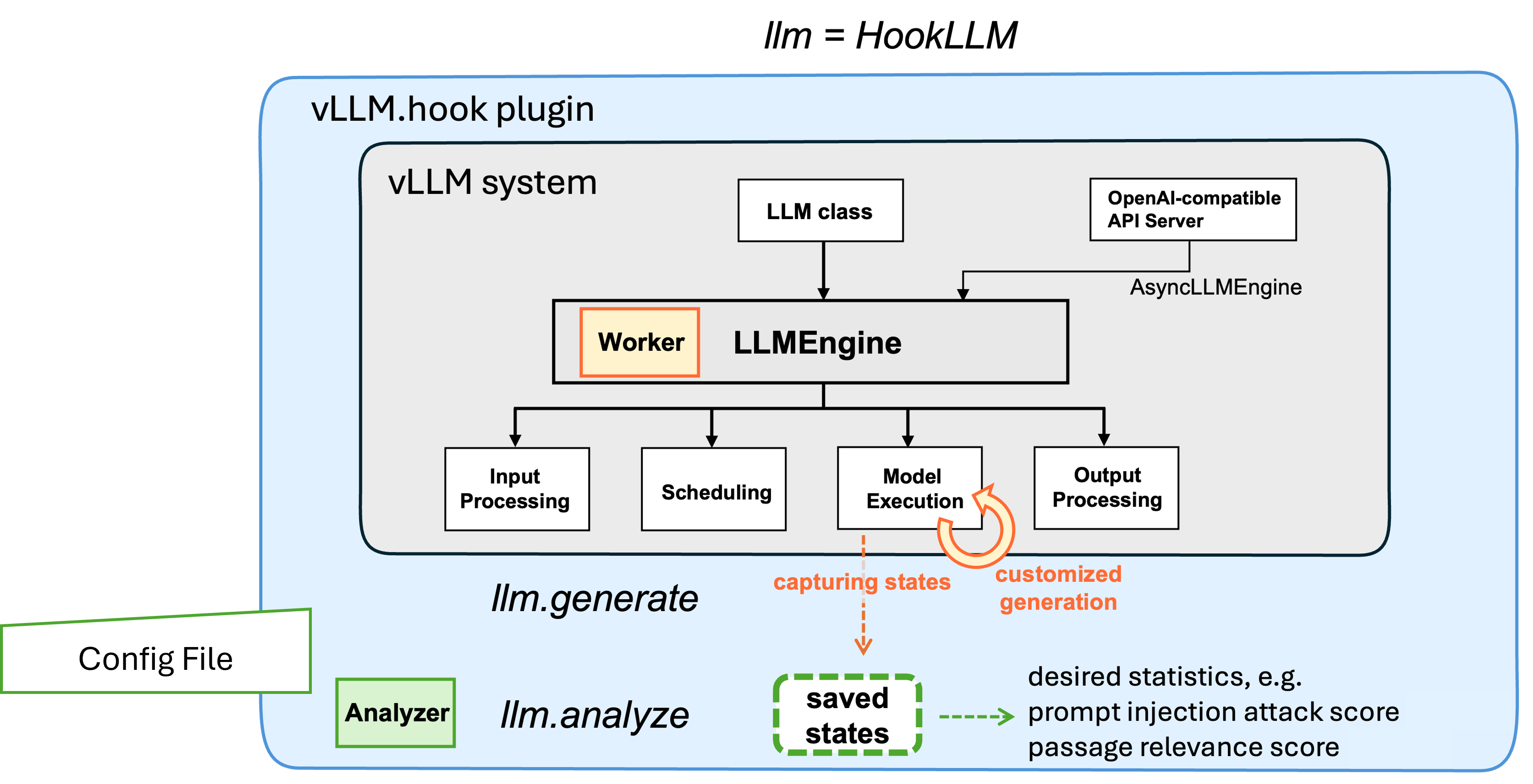}
    \caption{An illustration of vLLM-Hook orchestration. vLLM-Hook is a lightweight wrapper around a native vLLM system and uses \texttt{llm=HookLLM} (v0) or \texttt{llm=MiaLLM} (v1, vLLM-Hook MIA) to initialize an LLM instance, \texttt{llm.generate} for response generation (model forward pass), and \texttt{llm.analyze} for optional in-model analysis.}
    \label{fig:vLLM_Hook_internal}
\end{figure}


vLLM-Hook is built as a modular plugin library that enables users to capture internal model signals (passive programming), intervene during inference (active programming), and optionally analyze the captured signals. We give an illustration of vLLM-Hook orchestration in Figure~\ref{fig:vLLM_Hook_internal}. The core of our framework consists of two abstractions: \worker{} and \analyzer{}. On a higher level, \worker{} defines when and how we program within the vLLM pipeline, and \analyzer{} defines the evaluation metrics based on the captured signals (if any). Together with \worker{} and \analyzer{}, a \texttt{Config File} is provided to specify the more granular behavior of the \worker{} and \analyzer{}. For instance, it could include the profile of the important layers and attention heads of a model, the steering vector of a desired model behavior, and the statistic reduction method (take average vs. weighted summation), etc. We give an example of a config file as follows:
\begin{verbatim}
{
    "model_info": {
        "name": "granite3-8b-attn",
        "model_id": "ibm-granite/granite-3.1-8b-instruct"
    },
    "params": {
        "important_heads": [[6, 9], [7, 20], [8, 1], [8, 13], [8, 14], [8, 15], 
        [10, 2], [10, 3], [10, 6], [10, 21], [11, 4], [11, 30], [11, 31], [12, 2], 
        [12, 28], [13, 8], [13, 9], [13, 12], [14, 15], [14, 16], [14, 19], [14, 27], 
        [15, 6], [15, 7], [15, 20], [15, 23], [16, 12], [16, 14], [16, 16], [17, 7], 
        [17, 11], [17, 15], [17, 19], [17, 21], [17, 25], [17, 26], [18, 9], 
        [18, 17], [18, 20], [18, 28], [19, 1]]
    },
    "hookq":{
        "hookq_mode": "last_token"
    }
}
\end{verbatim}
In the above example, the config file describes the generating LM under \texttt{"model\_info"}, specifies the `[layer, head]' index of the important heads for the model analysis under \texttt{"params"/"important\_heads"}, and indicates we only need the query cache of the very last token through \texttt{"hookq"/"hookq\_mode"}. Other workers read their own respective variables, e.g. the hidden-state worker reads \texttt{"hidden\_states"} (which layers and token positions to capture) and the steering worker reads \texttt{"steering"} (steering method, coefficient, layers, and vector path). These components are managed through a lightweight registry and a \textbf{MiaLLM} wrapper class (\textbf{HookLLM} in v0), with \textbf{MiaClient} as its counterpart for \texttt{vllm serve}.

During the runtime, once the user calls \texttt{llm.generate} in active programming mode, \worker{} actively engages in the model execution phase and performs customized generation (e.g., model steering). In passive programming mode, \worker{} follows base generation flows and optionally capture internal states (e.g., attentions) during forward passes. After the forward pass has completed, the user can call \texttt{llm.analyze} and the \analyzer{} will assemble the saved states and compute the desired statistics such as prompt injection attack score, passage relevance score, etc.

\subsection{Worker}
Workers are integrated directly into the vLLM runtime and record/modify internal states during inference. 
In version 1, a worker is a mixin class that vLLM injects into its GPU worker through the \texttt{worker\_extension\_cls} mechanism, rather than a subclass of the GPU worker. vLLM-Hook MIA ships three workers: hidden-state capture, Q/K capture, and activation steering. By default, each worker bakes its capture or steering operation into the engine's CUDA graphs. A worker can also install PyTorch forward hooks on selected modules, which run only in eager mode (In version 0, a worker subclassed the standard vLLM GPU worker and overrode $load\_model$). We give a simplified example of the Q/K capture worker as follows:
\begin{verbatim}
class QKCaptureWorker:
    """Mixin injected into vLLM's GPU Worker via worker_extension_cls."""

    _hooks_installed: bool = False
    _captured_states: dict = {}

    def install_hooks(self):
        """Install forward hooks on all target attention modules."""
        if self._hooks_installed:
            return
        self._hooks_installed = True
        runner = self.model_runner
        require_v2_runner(runner)
        # stash each request's extra_args (e.g., which layers to capture)
        stash = install_request_arg_stash(runner)
        # ... wrap runner.prepare_inputs to record the current step
        model = runner.model
        # ... resolve model geometry and tensor-parallel shard

        def qkv_hook(input, module_name, attn_module=None):
            # ... for each request in the batch whose extra_args
            # contain "output_qk", slice its q/k tensors for the
            # requested layers and store them in self._captured_states
            ...

        # register hooks on attention modules
        for name, module in model.named_modules():
            layer_num = match_attn(name)
            if layer_num is not None:
                module.register_forward_hook(
                    lambda m, i, o, n=name: qkv_hook(i, n, m))

    def get_captured_states(self, external_req_id):
        # ... serialize one request's captured states
        ...

    def graph_install(self):
        """Install the CUDA-graph QK capture path."""
        # CUDA-graph path (default): instead of eager forward hooks,
        # allocate static per-layer capture buffers and wrap the runner's
        # prepare_inputs / execute_model so capture is routed and drained
        install_qk_hosts(self)
        install_execute_model_wrapper(self.model_runner, self)
\end{verbatim}

\subsection{Analyzer}
Analyzers are designed to (optionally) conduct analysis based on the saved statistics from the worker orchestration. Examples of analyzers include safety guardrails such as monitoring prompt injection attacks, and attention-based re-rankers that leverage selective attention weights to rank document relevance. 
Typically, an analyzer starts by retrieving the captured states, either passed in memory (\texttt{probes}, returned together with the generation output) or loaded from disk by run id. 
Then, it re-assembles the desirable statistics accordingly, e.g. selective attention weights. In the following, we show an example analyzer that uses saved keys and queries to recompute selective attentions for prompt injection attack risk evaluations:
\begin{verbatim}
class AttntrackerAnalyzer:
    def __init__(self, hook_dir: str, layer_to_heads: Dict[int, list]):
        self.hook_dir = hook_dir
        self.layer_to_heads = layer_to_heads

    def analyze(
        self,
        analyzer_spec: Optional[Dict] = None,
        run_id: Optional[str] = None,
        probes: Optional[Dict] = None,
    ) -> Optional[Dict]:
        # re-assemble selective attention weights from the captured Q/K,
        # passed in memory (probes) or loaded from disk (run_id)
        attention_weights = self.compute_attention_from_qk(run_id, probes=probes)

        # evaluate prompt injection attack risks
        score = self.attn2score(attention_weights, analyzer_spec['input_range'],
                analyzer_spec['attn_func'])

        return {"score": score}

    def compute_attention_from_qk(self, run_id=None, probes=None) -> Dict[str, Dict]:
        if probes is not None:
            qk_cache = probes["qk_cache"]
        else:
            qk_cache = load_and_merge_qk_cache(self.hook_dir, run_id)["qk_cache"]
        # ... recompute the configured heads' attention from the saved q/k
        ...

    def attn2score(self, batch_attention: List[Dict[str, Dict]],
        batch_input_range: List[Tuple[Tuple[int, int], Tuple[int, int]]],
        attn_func: str = "sum_normalize"
    ) -> List[float]:
        # ... following Attention Tracker and calculate the aggregated attention on
        # the instruction vs the attention over both the instruction and user query
        ...
\end{verbatim}

\section{\textbf{vLLM Hook} Use Cases}
\label{sec:demos}

vLLM-Hook is an extensible framework that aims to allow selective access to model internals during the inference. We provide demonstrations of vLLM-Hook through a series of example scripts in the \texttt{examples/} folder of the repository (\url{https://github.com/IBM/vLLM-Hook/tree/main/examples}). We list the use cases in chronological order below:
\begin{itemize}
    \item \textbf{In-model monitoring of prompt injection attacks}
    \begin{itemize}
        \item TL;DR: Attention Tracker~\cite{hung2025attention} monitors prompt injection attacks via the aggregated attention scores of the important heads on the instruction prompt, also called focus score. Low focus score indicates potential malicious queries.
        \item Hook: the \texttt{capture\_qk} worker saves the query and key tensors of the configured heads, and the \texttt{attn\_tracker} analyzer returns one focus score per prompt.
        \item Demo: \texttt{examples/demo\_attntracker.py}
    \end{itemize}
    \item \textbf{Attention-based document ranking}
    \begin{itemize}
        \item TL;DR: Core reranker~\cite{tran2025contrastive} is an attention-based reranker that leverages attention weights from selected transformer heads to produce document relevance scores.
        \item Hook: the \texttt{capture\_qk} worker saves the query and key tensors, and the \texttt{core\_reranker} analyzer returns relevance scores and a ranking for each case.
        \item Demo: \texttt{examples/demo\_corer.py}
    \end{itemize}
    \item \textbf{Model steering via activation steering}
    \begin{itemize}
        \item TL;DR: Activation steering allows you to bias the model's behavior by nudging internal activations in specific directions. In~\cite{stolfo2025improving}, authors focus on instruction following capability and compute the steering vectors as the difference in activations between inputs with and without instructions.
        \item Hook: the \texttt{steer} worker modifies the hidden states of the configured decoder layers during generation, and the steering method, coefficient, layers and vector are given in the config and can be overridden per request. This use case is active programming and needs no analyzer.
        \item Demo: \texttt{examples/demo\_actsteer.py}
    \end{itemize}
    \item \textbf{Hidden-state probe}
    \begin{itemize}
        \item TL;DR: A demonstration of the basic building block for probing methods that capture the hidden states of selected layers for each prompt.
        \item Hook: the \texttt{capture\_hs} worker saves the hidden states, and the \texttt{hidden\_states} analyzer returns them per layer, either as they are or reduced by mean or norm.
        \item Demo: \texttt{examples/demo\_hiddenstate.py}
    \end{itemize}
    \item \textbf{Science hallucination detection through activation probe}
    \begin{itemize}
        \item TL;DR: An activation probe on the captured hidden states labels answers to scientific claims~\cite{cao2025detecting} as entailment, contradiction or unverifiable.
        \item Hook: the \texttt{capture\_hs} worker saves the hidden state at the probe layer, and the \texttt{science\_hallucination} analyzer returns a prediction and a label for each case.
        \item Demo: \texttt{examples/demo\_scihal.py}
    \end{itemize}
    \item \textbf{H-Node hallucination detection through activation probe}
    \begin{itemize}
        \item TL;DR: An activation probe on the last-token hidden state of one layer assigns a hallucination probability to each prompt before the first output token is sampled~\cite{yocam2026h}.
        \item Hook: the \texttt{capture\_hs} worker saves the hidden state at the probe layer, and the \texttt{hnode\_hallucination} analyzer returns probabilities and grounded or hallucinated verdicts.
        \item Demo: \texttt{examples/demo\_halludetect.py}
    \end{itemize}
    \item \textbf{Attention-based schema linking for text-to-SQL}
    \begin{itemize}
        \item TL;DR: AttnLink-U~\cite{song2026attnlink} ranks the database columns needed to answer a question from the model's attention at the generation-start position, in a single prefill pass and without task-specific training or generating SQL. A set of columns is then selected by cumulative probability mass.
        \item Hook: the \texttt{capture\_qk} worker saves the attention of a single configured head, and the \texttt{attnlink} analyzer returns scores, a ranking and the selected set.
        \item Demo: \texttt{examples/demo\_attnlink.py}
    \end{itemize}
    \item \textbf{Tool-call risk assessment for agents through activation probe}
    \begin{itemize}
        \item TL;DR: An activation probe on the last prompt token's hidden state scores an agent step before the model acts, estimating the probability that the next action follows a prompt injection that arrived in a tool result.
        \item Hook: the \texttt{capture\_hs} worker saves the hidden state at the probe layer, and the \texttt{tool\_call\_risk} analyzer returns probabilities and pass or flag verdicts.
        \item Demo: \texttt{examples/demo\_toolcallrisk.py}
    \end{itemize}
\end{itemize}

\section{How to Contribute to \textbf{vLLM Hook}?}
We envision vLLM Hook as an organic innovation engine that will enable the development and deployment of advanced techniques for monitoring and adjusting vLLM models. These techniques require the use of programming internal states during inference.   

We invite the community to contribute to the vLLM Hook project to expand its core functions, use cases, and demonstrations. Significant contributors will be invited to co-author future vLLM technical reports.
\paragraph{To contribute,}
\begin{enumerate}
    \item \textbf{Fork} this repository: \url{https://github.com/IBM/vLLM-Hook/tree/main}
    \item \textbf{Create a branch} (\texttt{git checkout -b feature/amazing-feature})  
    \item \textbf{Commit} your changes (\texttt{git commit -m `Add amazing feature'})  
    \item  \textbf{Push} to your branch (\texttt{git push origin feature/amazing-feature})  
    \item \textbf{Open a Pull Request and request approval from the admin}  
\end{enumerate}
\paragraph{Guidelines}
\begin{itemize}
    \item Users are encouraged to define new worker/analyzer. Please discuss with the maintainers before modifying \texttt{mia/llm.py}, \texttt{mia/client.py}, \texttt{mia/core/\_plugin.py}, or anything under \texttt{mia/core/}.
    \item Include examples and documentation for new features.
    \item New use cases must be added to \texttt{docs/use\_cases/README.md}, together with the contributor's GitHub handle.
\end{itemize}


\section{Acknowledgment}
We thank David Cox for his comments on ``Do your techniques work on vLLM?'' This package was created as a result. We thank Kush Varshney's support and feedback. We thank Hendrik Strobelt and Enara Vijil for their feedback on the implementation and testing. We also thank all community contributors for the use cases, workers, and analyzers they added to the repository.

\bibliographystyle{alpha}
\bibliography{sample}

\end{document}